\documentclass[11pt,a4paper]{article}
\usepackage[hyperref]{acl2020}
\usepackage{times}
\usepackage{latexsym}
\usepackage{booktabs}
\usepackage[position=bottom]{subfig}
 \usepackage{amssymb}
 \usepackage{amsmath}
\usepackage[colorinlistoftodos,prependcaption,textsize=small]{todonotes}

\usepackage{pbox}

\usepackage[title]{appendix}
\usepackage{placeins}

\newenvironment{itemize*}%
 {\begin{itemize}%
  \setlength{\itemsep}{0pt}%
  \setlength{\parskip}{0pt}}%
 {\end{itemize}}
 \newenvironment{enumerate*}%
 {\begin{enumerate}%
  \setlength{\itemsep}{0pt}%
  \setlength{\parskip}{0pt}}%
 {\end{enumerate}}
\usepackage{microtype}

\aclfinalcopy 
\def\aclpaperid{2819} 

\definecolor{myblue}{rgb}{0.9, 0.1, 0.94}
\definecolor{mygreen}{rgb}{0.64, 0.76, 0.68}
\definecolor{myyellow}{rgb}{0.98, 0.94, 0.75}
\definecolor{mygreen}{rgb}{0.68, 0.9, 0.6}

\definecolor{cadred}{rgb}{0.89, 0.0, 0.13}

\definecolor{blush}{rgb}{0.87, 0.36, 0.51}
\definecolor{darkpastelpurple}{rgb}{0.59, 0.44, 0.84}
\definecolor{orangepeel}{rgb}{1.0, 0.62, 0.0}
\definecolor{teal}{rgb}{0.0, 0.5, 0.5}
\definecolor{vividburgundy}{rgb}{0.62, 0.11, 0.81}

\title{How well do you know your summarization datasets?}

\author{Priyam Tejaswin \thanks{\ \ This author was the primary contributor.} \qquad Dhruv Naik\qquad Pengfei Liu \thanks{\ \  Corresponding author.}\\
  Language Technologies Institute, Carnegie Mellon University, Pittsburgh, PA\\
  \texttt{\{ptejaswi, drn, pliu3\}@andrew.cmu.edu}
  }

\begin{document}
\maketitle
\begin{abstract}

State-of-the-art summarization systems are trained and evaluated on massive datasets scraped from the web. Despite their prevalence, we know very little about the underlying characteristics (data noise, summarization complexity, etc.) of these datasets, and how these affect system performance and the reliability of automatic metrics like ROUGE. In this study, we manually analyse 600 samples from three popular summarization datasets. Our study is driven by a six-class typology which captures different noise types (missing facts, entities) and degrees of summarization difficulty (extractive, abstractive). We follow with a thorough analysis of 27 state-of-the-art summarization models and 5 popular metrics, and report our key insights: (1) Datasets have distinct data quality and complexity distributions, which can be traced back to their collection process. (2) The performance of models and reliability of metrics is dependent on sample complexity. (3) Faithful summaries often receive low scores because of the poor diversity of references. We release the code, annotated data and model outputs.\footnote{ \href{https://github.com/priyamtejaswin/howwelldoyouknow}{https://github.com/priyamtejaswin/howwelldoyouknow}}


\end{abstract}

\section{Introduction}
\label{section:intro}

The past few years have witnessed major breakthroughs and improvements in automatic summarization \cite{see-etal-2017-get, celikyilmaz2018deep,jadhav2018extractive,liu2019text,liu2019fine,dou2020gsum,yuan2021can,liu2020refactor}. Apart from the improvements in the summarization model architectures \cite{zhang2019pegasus,zhong2020extractive}, this growth has been aided by large-scale datasets \cite{nallapati2016abstractive,narayan2018don,sharma2019bigpatent} and automatic evaluation metrics \cite{lin2004rouge,zhao-etal-2019-moverscore,kryscinski-etal-2020-evaluating} which are used for tuning hyperparameters and comparing models. While the reliability of these metrics has been explored extensively \cite{peyrard-2019-studying, bhandari-etal-2020-evaluating,fabbri2020summeval}, few studies have focused on the underlying characteristics of different datasets, and how these impact model performance and metric reliability.

Datasets like CNN/DailyMail \cite{nallapati2016abstractive}, Gigaword \cite{rush-etal-2015-neural}, XSum \cite{narayan2018don}, and many more \cite{wang-ling-2016-neural, koupaee2018wikihow, Kim-2019-NAACLHLT, ganesan-etal-2010-opinosis}  were collected by scraping a large collection of web-pages. And for all the benefits this approach offers (seemingly infinite samples, diverse subjects, etc) there are some caveats:

\paragraph{Data Noise} We have no idea about the \textbf{noise} in the dataset. In the context of text summarization, noise could be an incomplete or irrelevant reference. At the moment, its quantity and impact on the performance is unknown.

\paragraph{Summarization Complexity}
What do we really know about the nature of samples in the dataset? Gigaword is a headline generation dataset with short sources and references. Does this imply a higher volume of simpler (i.e. more \textit{extractive}) samples? The degree of summarization complexity, and its impact on model performance is unknown.

Exploring these open questions is critical for two reasons: (1) Information about the noise could lead to more informed data collection and pre-processing methods: in a recent study, \citet{kryscinski-etal-2019-neural} quantified HTML artefacts in popular summarization datasets, and proposed ways to detect and remove them. (2) Awareness about the complexity could better explain model performance, metrics, and even lead to new model architectures. In the tasks of machine comprehension and question answering, \citet{chen-etal-2016-thorough} and \citet{yatskar-2019-qualitative} manually inspected random samples and drew insights which led to new state-of-the-art models. Such analysis could also help researchers choose datasets and metrics more carefully.

In this study, we perform intrinsic and model-centric evaluation of three popular summarization datasets (Gigaword, CNN/DM and XSum). We are interested in answering the following questions: 
\paragraph{Q1. What are the underlying intrinsic properties of summarization datasets?}
We are interested in (1) Identifying and quantifying the different types of ``noise" that could occur and could penalize models. (2) Whether samples have different levels of difficulty. Armed with this, we ask the following questions.

\paragraph{Q2 a. How do these properties impact model performance?} Specifically, we'd like to know (1) If, and how, the performance varies across the different types of samples discovered from Q1. (2) If the performance is consistent across metrics.

\paragraph{Q2 b. If the reliability of metrics changes with these properties?} This is motivated (in part) from prior metric-analysis studies, where researchers have explored inter-metric agreement and alignment with human-judgement under different conditions \cite{peyrard-2019-studying, bhandari-etal-2020-evaluating}. Here we are more interested in knowing if the metrics are more correlated with human judgement for simpler samples, than complex ones.

\paragraph{}Large-scale automatic intrinsic dataset evaluation has been explored with some promising results \cite{bommasani-cardie-2020-intrinsic}. However, these methods rely on heuristics like content-value, density and compression \cite{grusky-etal-2018-newsroom}. We are interested in a more fine-grained, interpretable analysis that can only come from manual inspection, much like the analysis by \citet{chen-etal-2016-thorough} and by \citet{yatskar-2019-qualitative}. To that end, we first define a six-class typology: the first three classes cover types of data-noise and the last three cover varying degrees of summarization difficulty. We then proceed to answer the aforementioned research questions, and discuss our key observations which are summarized below: 

\paragraph{Key Observations:} (1) Datasets have distinct modalities -- a mix of simpler samples (which we call \textit{Extractive}) and complex ones (which we call \textit{Paraphrase} and \textit{Inference}. (2) Gigaword is \textit{majorly} Extractive but suffers from data noise (45\% of the targets have some key entity, or fact that is absent from the source). (3) CNN/DM is relatively cleaner, and the authors' attempts to create a more abstractive dataset seems to be successful compared with Gigaword (only 18\% of samples are Extractive). (4) XSum has no Extractive samples, but also has the greatest fraction of noise: 54\% of the test samples have key entities or facts missing from the source. (5) Within the datasets, the broad performance trends between the typology classes are \textit{consistent across all metrics}: \textbf{simpler samples score higher than complex ones}. (6) Metric reliability is also complexity dependent: On CNN/DM the agreement with human judgement decreases as summarization complexity increases.

The remainder of the paper is organised as follows: in Section \ref{section:dataset} we answer \textbf{Q1}, describe the three datasets, define the typology, and present results from the annotation. In Section \ref{section:q2a} we explore \textbf{Q2 a.} and evaluate different models on a variety of metrics (automatic and human-judgement). In Section \ref{section:q2b} we explore \textbf{Q2 b.} and investigate metric reliability. In Section \ref{section:discussion} we share some learnings from our experience. We conclude with Section \ref{section:conclusion}.


\section{Evaluating the intrinsic properties of summarization datasets (Q1)}
\label{section:dataset}

\begin{table}[htbp]
  \centering
  \scriptsize
    \begin{tabular}{lrrrrrr}
    \toprule
          & \multicolumn{2}{c}{\textbf{Length(Doc)}} & \multicolumn{2}{c}{\textbf{Length(Ref)}} & \multicolumn{2}{c}{\textbf{Sample}} \\
\cmidrule{2-7}          & \multicolumn{1}{c}{train} & \multicolumn{1}{c}{test} & \multicolumn{1}{c}{train} & \multicolumn{1}{c}{test} & \multicolumn{1}{c}{train} & \multicolumn{1}{c}{test} \\
    \midrule
    Gigawords &    31   &  29     & 8      &  8     &  3.8M     & 1.9K \\
    CNNDM &   691    &  682     &   51    &  54     &    287K   & 11K \\
    XSum  &  374  &  376  &   21  &  21  &  204K  &  11.3K \\
    \bottomrule
    \end{tabular}%
      \caption{Statistics of the three datasets. Length refers to the average number of words per Document/Reference.}
  \label{tab:data_stats}%
\end{table}%

\subsection{Datasets for Annotation}

Among many summarization datasets, we choose the following:

\begin{table*}[ht]
  \tiny

    \begin{tabular}{ p{2cm}  p{12cm} }
    \toprule
    Label & Source \\ \cmidrule{2-2} 
    Dataset & Target \\ \cmidrule{2-2} 
          & SoTA \\ \midrule
    
    Incomplete / Irrelevant  & 
    Andre Blom and Mark Scharrenberg scored tries and some tactical kicks in the final 10 minutes sent the United States to the Rugby World Cup with a 21-16 victory over Uruguay on Saturday . \\ \cmidrule{2-2}
    Gigaword & \textcolor{red}{London testing , please ignore} . \\ \cmidrule{2-2}
    & United States beats Uruguay 21-16 in Rugby World Cup .
     \\
    \midrule
    
    Entity Missing &
    \textcolor{orangepeel}{The United States claimed credit} Tuesday for a \textcolor{blush}{ceasefire that ended fighting between Israel and Lebanese} guerrillas , and rejected suggestions that it was forced to model the agreement after a French draft .
    \\ \cmidrule{2-2}
    Gigaword & \textcolor{orangepeel}{US takes the credit} for \textcolor{blush}{Israel-}\textcolor{red}{Hezbollah} \textcolor{blush}{ceasefire by} \textcolor{red}{Carole Landry} . \\ \cmidrule{2-2}
    & Us claims credit for lebanon ceasefire .
    \\
     \midrule
     
    Extractive  &
    
    Ed Miliband’s US adviser pays no tax in Britain on his reported £300,000 salary, he has admitted. \textcolor{blush}{David Axelrod masterminded two presidential election victories for Barack Obama} and was \textcolor{darkpastelpurple}{hired by the Labour leader amid great fanfare last year}. He has helped refine Mr Miliband’s message ...\textit{(truncated)} ... have been aware of Labour’s eye-catching crackdown on non-doms last week. \textcolor{teal}{But speaking in the US where he is promoting his autobiography, Mr Axelrod revealed he is not resident for tax purposes in the UK.} Asked whether he pays tax in Britain, he told the Daily Telegraph: ‘I don’t do my accounting so I don’t know but I’m not in residence there.’ \textcolor{orangepeel}{Labour confirmed it pays Mr Axelrod in dollars through his consultancy firm} and that he ‘lives in the US, works in the US and pays taxes in the US’. ... (\textit{truncated}) 
    \\ \cmidrule{2-2}
    
    CNN-DM & \textcolor{blush}{David Axelrod masterminded two Obama presidential election victories} . \textcolor{darkpastelpurple}{He was hired by Labour leader Ed Miliband amid great fanfare last year} . \textcolor{teal}{Revealed at a book launch that he is not resident for tax purposes in UK} . \textcolor{orangepeel}{Labour confirms it pays Mr Axelrod in dollars through consultancy firm} . \\ \cmidrule{2-2}
    
    & David Axelrod masterminded two presidential election victories for Barack Obama .  He was hired by the Labour leader amid great fanfare last year .  Has helped refine Mr Miliband ’s message about tackling the cost of living and making sure the wealthy pay their fair share .  Mr Axelrod makes infrequent visits to the UK to meet Mr Miliband and offers advice by phone .
    
    \\
    \midrule
    
    Paraphrase &
    \textcolor{blush}{The number of women in Britain becoming nuns is at a 25-year high. Figures from the Catholic Church show the number of women taking Holy
    Vows has trebled from 15 in 2009 to 45 last year}. \textcolor{darkpastelpurple}{From a low of only seven in 2004}, the figure has been rising for the past decade. Theodora Hawksley, 29, was until recently a post-doctoral researcher in theology at the University of Edinburgh. But at the beginning of the year she decided to become a nun. \textit{(truncated)}. Far from being trapped in traditional habits, Miss Hawksley said her order tends to dress down in T-shirts and jeans. \textcolor{teal}{Father Christopher Jamison, director of the National Office for Vocation of England and Wales, said: ‘There is a gap in the market} for meaning in our culture. \textcolor{teal}{One of the ways women may find that meaning is through religious life.’} Sister Cathy Jones, religious life vocations promoter at the office, said: \textit{(truncated)} . \\ \cmidrule{2-2}

    CNN-DM & \textcolor{blush}{Figures from the Catholic Church show more and more becoming nuns}. \textcolor{darkpastelpurple}{The number of women taking Holy Vows stood at just seven back in 2004} . But that figure had \textcolor{blush}{risen to 15 in 2009 and increased further to 45 last year} . \textcolor{teal}{One father said a ’ gap in the market for meaning ’ led people toward religion} . \\ \cmidrule{2-2}
    
    & Figures from Catholic Church show number of women taking Holy Vows has trebled from 15 in 2009 to 45 last year . From a low of seven in
    2004 , the figure has been rising for the past decade . Theodora Hawksley , 29 , was until recently a post - doctoral researcher in theology at
    the University of Edinburgh . But at the beginning of the year she decided to become a nun .
    
    \\
    \midrule
    
    Inference  & \textcolor{orangepeel}{Three Malaysian and Indonesian seamen kidnapped by Philippine Abu Sayyaf kidnap-for-ransom group} \textcolor{darkpastelpurple}{allegedly had been executed} and the
    skeletons discovered in the southern Philippines are believed to be their remains , \textcolor{teal}{a local television reported} Wednesday . \\ \cmidrule{2-2}
    
    Gigaword & \textcolor{orangepeel}{Abu Sayyaf hostages} \textcolor{darkpastelpurple}{allegedly executed} : \textcolor{teal}{report} . \\ \cmidrule{2-2}
    
    & 3 filipino , Indonesian seamen executed in southern Philippines . \\
    
    \bottomrule
    
    \end{tabular}%
    
      \caption{Examples for each of the six categories. Text spans with the same colors correspond to the same fact in the source and target. Target spans in \textcolor{red}{RED} are missing or unsupported in the source. The last sample is ``Inference" because the writer will have to understand the concept of hostages, and then generalise from the group to an individual.
      }
  \label{tab:datasamples}%
\end{table*}%

\noindent \textbf{Gigaword} is a summarizaiton dataset extracted from news articles \cite{rush-etal-2015-neural}\footnote{We use the version most commonly used by summarization systems: \href{https://github.com/harvardnlp/sent-summary}{https://github.com/harvardnlp/sent-summary}}.

\noindent \textbf{CNN/DailyMail} or ``CNN/DM" question answering dataset \cite{hermann2015teaching, nallapati2016abstractive} is commonly used for summarization. The dataset consists of online news articles paired with human-generated summaries.\footnote{We use the non-anonymized data as \citet{see-etal-2017-get}.}

\noindent \textbf{XSum} or ``Extreme Summarization" \cite{narayan2018don} was constructed from online news articles for highly abstractive summarization.

We consider these datasets because of their popularity, and the difference in the nature of samples. The latter enables a more comprehensive analysis; Table \ref{tab:data_stats} captures the size of source and target documents along with the number of samples.

\subsection{Typology Definition}
The classes are defined below in order of priority. Some examples are in Table \ref{tab:datasamples}. Readers may refer to the Appendix \ref{section:appx-giga}, \ref{section:appx-cnndm}, \ref{section:appx-xsum} for more examples.
\begin{itemize*}
    \item \textbf{Incomplete/Irrelevant}: The target summary ends abruptly. Or the source and target are unrelated.
    \item \textbf{Entity Missing}: The target summary contains entities (names, dates, events, etc) that are absent from the source.
    \item \textbf{Evidence Missing}: The target summary is based on concepts which are absent from the source. However, the target is not \textbf{Incomplete} and all \textbf{Entities} are present.
    \item \textbf{Extractive}: The target is constructed by copying tokens from the source, mostly in-order of their appearance. Minor modifications, like stemming and abbreviating, are permitted. Word substitutions, and additions, are limited to a few. No reasoning, conclusion or co-ref resolution is performed as part of the summarization. The complete context of the target should be present in the source.
    \item \textbf{Paraphrase}: The majority of tokens in the target are substituted, or appear out of order, or both. There is no reasoning, conclusion or co-ref resolution. The complete context of the target should be present in the source.
    \item \textbf{Inference}: A non-trivial ``inference" activity has to be completed to construct the target: some reasoning, conclusion, or complex co-reference resolution. The complete context of the target should be present in the source.
\end{itemize*}

We annotate 200 samples from each dataset, on par with similar studies on intrinsic evaluation \cite{chen-etal-2016-thorough, cao2017faithful}. Two authors annotate samples independently. Annotations matched for 70\%, 68\% and 73\% of Gigaword, CNN-DM and XSum samples, respectively. Disagreements were discussed between all authors before arriving at a consensus for the final label.

\subsubsection{Motivation and Advantages}
\label{section:motivate}
To the best of our knowledge, summarization datasets have not been manually analysed in this manner. A review of the most relevant summarization dataset analysis research shows that the most common form of intrinsic evaluation is to use surface-level heuristics. Most studies only cover a part of our typology, while almost all studies ignore the noise present in datasets. 

\paragraph{Coverage , Density, Redundancy} \citet{grusky-etal-2018-newsroom, bommasani-cardie-2020-intrinsic, zhong-etal-2019-closer} use similar forms of token-level coverage between the source and the reference to measure the \textit{extractiveness} of the summary. In it's simplest form, this is a ratio of the number of overlapping tokens and reference length. In our definition of \textbf{Extractive}, we first set a meaninful, well-defined criterion, and then manually check for \textit{extractive} references, while allowing for some relaxations.

\paragraph{Content Compression} In most papers \cite{grusky-etal-2018-newsroom, zhong-etal-2019-closer, bommasani-cardie-2020-intrinsic}, the summarization complexity is defined by a compression ratio (usually the normalized word-count ratio of the source and reference). As a standalone metric, this does indeed capture the difficulty in replication. However, token rearrangement, substitution, reformulation is ignored in this measure of ``complexity". To combat this, we distinctly defined \textbf{Paraphrase} and \textbf{Inference}. By manually analysing samples, we are able to differentiate between the obviously simple Extractive samples, the relatively tougher Paraphrase samples and the most difficult Inference samples. Together these three offer a highly intuitive classification of samples. Part of the reason that the Machine Comprehension analysis by \citet{chen-etal-2016-thorough} was so effective was the interpretability of their classes. We hope our analysis will also enable researchers to improve summarization models.

\paragraph{Noise} Prior works have not focused on quantify the noise in popular datasets. Moreover, none of these metrics are designed to account for noise or factual inconsistencies. A high value for content compression might imply a high-degree of summarization complexity. But this ignores the possibility that the source-reference pair is unrelated (like row 1 in Table \ref{tab:datasamples}). In addition, the manual analysis allows us to identify factual errors and co-ref errors.

\paragraph{}This is not to say the typology is perfect and exhaustive. Limitations and possible extensions to our typology are discussed in Section \ref{section:discussion}.

\subsection{Dataset Analysis}

\begin{figure}[t]
 \centering
 \includegraphics[width=1\columnwidth]{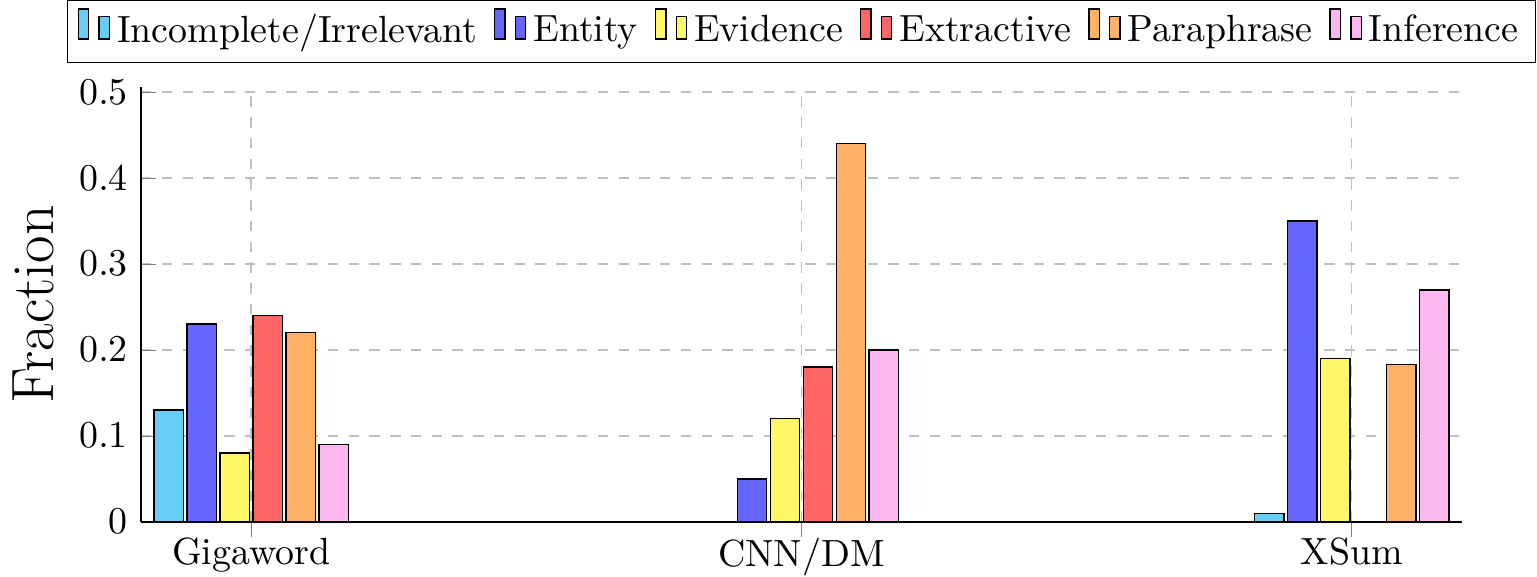}
 \caption{Distribution of the different class of samples in all datasets.}
 \label{fig:datasets}
\end{figure}

The distribution of classes in the datasets is in Figure \ref{fig:datasets}. We have made the following \textbf{key observations} in our analysis of the labels.

\paragraph{Gigawords is Extractive, but very noisy.} 24.5\% of summaries are Extractive, but 44.5\% of samples belong to Entity Missing, Evidence Missing, or Incomplete. Not unexpected considering the ``headline" nature of the samples.

\paragraph{XSum is Abstractive, but also very noisy.}  The authors \citep{narayan2018don} designed the dataset to be highly abstractive. This is reflected in the distribution: there were no Extractive samples in our analysis, suggesting a significantly higher level of difficulty. However, 55\% of samples belong to Entity Missing, Evidence Missing, or Incomplete classes. The remaining 45\% belongs to Paraphrase and Inference categories. Since we found only two incomplete samples, this class is ignored in all further XSum analysis.

\paragraph{CNN/DM is cleaner, and lives up to the design goals.} The authors \citep{hermann2015teaching} designed CNN/DM to be \textit{abstractive} in nature, and this is reflected in the distribution: 64\% of samples belong to Paraphrase and Inference categories. Of the three, CNN/DM has the lowest fraction of factual and data noise: there are no Incomplete/Irrelavant samples, and only 18\% of samples belong to Entity Missing and Evidence Missing.

The degree with which missing facts affects automatic evaluation varies. In some samples, one or two entities are missing (like Row 2 in Table \ref{tab:datasamples}), but in others multiple facts are missing. Empirical analysis of model performance for each class of samples is discussed in Section \ref{section:q2a}.


\section{Performance on different classes (Q2 a)}
\label{section:q2a}
In this section, we list the different models and metrics considered for analysis, and then describe how model performance varies across class labels.

\subsection{Models for evaluation}

We collect outputs from 7 systems for Gigaword:
\noindent
(1) \textsc{Pegasus}~\citep{zhang2019pegasus}, 
(2) \textsc{Prophet}~\citep{qi-etal-2020-prophetnet} ~\citep{lewis-etal-2020-bart},
(3) \textsc{UniLM} ~\cite{dong2019unified} ,
(4) \textsc{Biset} ~\cite{DBLP:conf/aaai/SongWFL020},
(5) \textsc{ConCopy} ~\cite{wang-etal-2019-biset} ,
(6) \textsc{PointerGenerator} ~\cite{see-etal-2017-get},
(7) \textsc{PointerGeneratorCopying}~\cite{see-etal-2017-get} 

For CNN/DM, we use the outputs of 11 top-performing summarization systems collected by \citet{bhandari-etal-2020-evaluating}\footnote{https://github.com/neulab/REALSumm}:  
\noindent
(1) \textsc{HeterGraph}~\citep{wang2020heterogeneous}, 
(2) \textsc{MatchSumm} ~\citep{lewis-etal-2020-bart},
(3) \textsc{Refresh} ~\cite{narayan-etal-2018-ranking} ,
(4) \textsc{TwoStageRL} ~\cite{DBLP:conf/aaai/SongWFL020},
(5) \textsc{Neusumm}~\cite{wang-etal-2019-biset} ,
(6) \textsc{BottomUp}~\cite{gehrmann2018bottom}
(7) \textsc{SemSim}~\cite{yoon2020learning}
(8) \textsc{UniLM}~\cite{dong2019unified}
(9) \textsc{BartAbstractive}~\cite{lewis-etal-2020-bart} 
(10) \textsc{BanditSumm}~\cite{dong-etal-2018-banditsum}
(11) \textsc{BartExtractive}~\cite{lewis-etal-2020-bart}

For XSum, we use the outputs of 9 different summarization systems: 
\noindent
(1) \textsc{ConvSeq2Seq}~\citep{gehring-etal-2017-convolutional},
(2) \textsc{TConvS2S}~\citep{narayan2018don}
(3) \textsc{PointerGenerator}~\citep{see-etal-2017-get},
(4) \textsc{Bart}~\citep{lewis-etal-2020-bart},
(5) \textsc{PreSummExtractive}~\citep{liu2019text},
(6) \textsc{PreSummAbstracctive}~\citep{liu2019text},
(7) \textsc{PreSummTransformer}~\citep{liu2019text},
(8) \textsc{LEAD}~\citep{nenkova2005lead},
(9) \textsc{ExtOracle}~\citep{nallapati2017summarunner}

\subsection{Metrics for evaluation}
Existing summarization systems are usually evaluated using automated metrics or manually using human judgments. We list popular automatic metrics explored in this work. Except for the last two, all outputs from every model is scored on the following metrics.

\noindent\textbf{ROUGE-1/2/L} measure overlap of unigrams, bigrams and longest common subsequence.  respectively\footnote{\label{pyrouge}For ROUGE-1,2, and L, we used the Python implementation: \href{https://github.com/sebastianGehrmann/rouge-baselines}{https://github.com/sebastianGehrmann/rouge-baselines}} \citep{lin2004rouge}.


\noindent\textbf{BERTScore (BS)} measures soft overlap between contextual BERT embeddings of tokens between the two texts\footnote{Used code at \href{https://github.com/Tiiiger/bert_score}{github.com/Tiiiger/bert\_score}} \citep{bert-score}.

\noindent\textbf{MoverScore (MS)} applies a distance measure to contextualized BERT and ELMo word embeddings\footnote{We used a faster version of the code provided by the author at \href{https://github.com/AIPHES/emnlp19-moverscore}{github.com/AIPHES/emnlp19-moverscore}} \citep{zhao-etal-2019-moverscore}.

\noindent\textbf{FactCC}  is introduced to measure the fact consistency between the generated summaries and source documents \cite{kryscinski-etal-2020-evaluating}. 
Due to issues with the setup and training procedure, this metric was only used in the CNN/DM analysis.

\noindent\textbf{Human Pyramid (HP)} provides a robust technique for evaluating content selection by exhaustively obtaining a set of Semantic Content Units (SCUs) from a set of references, and then scoring system summaries on the number of SCUs that can be inferred \cite{nenkova-passonneau-2004-pyramid-og}. We use the scores shared by \citet{bhandari-etal-2020-evaluating} for the first 100 samples of CNN/DM subset.

\label{section:experiments}

\subsection{Model Performance}
\begin{figure}[t]
    \centering
    \includegraphics[width=1\columnwidth]{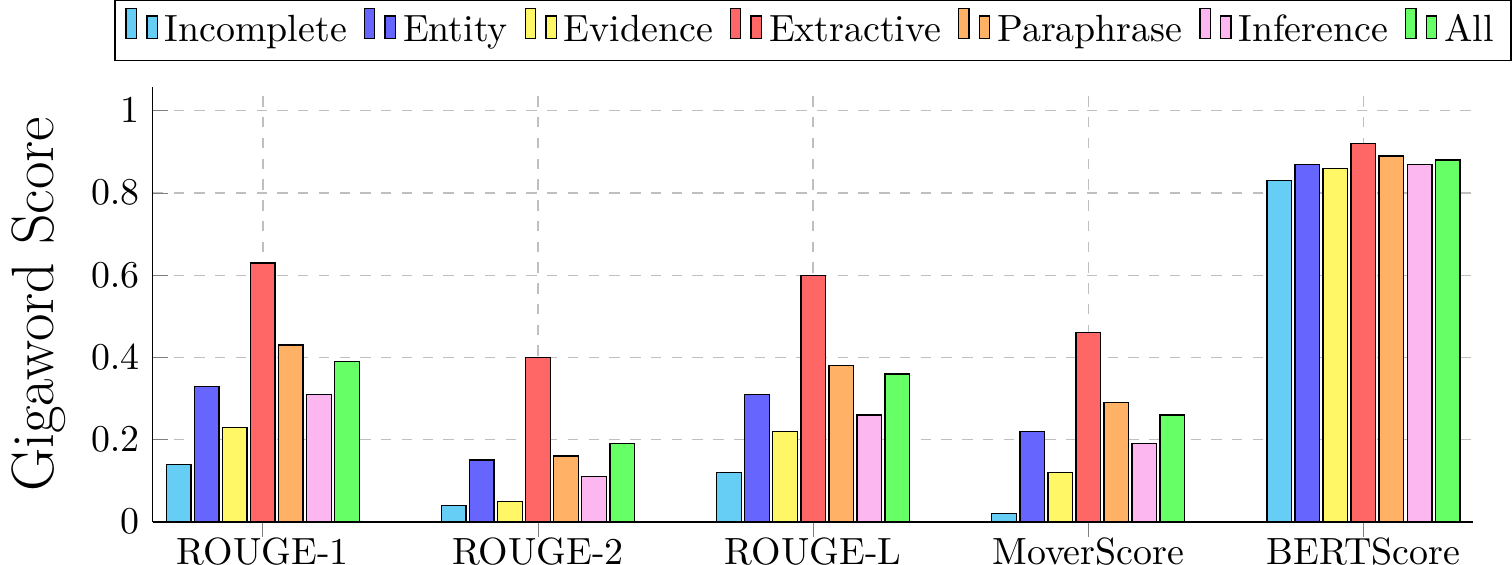}
    \caption{Gigaword class-level performance, averaged across all models.}
    \label{fig:classgiga}
\end{figure}

\begin{figure}[t]
    \centering
    \includegraphics[width=1\columnwidth]{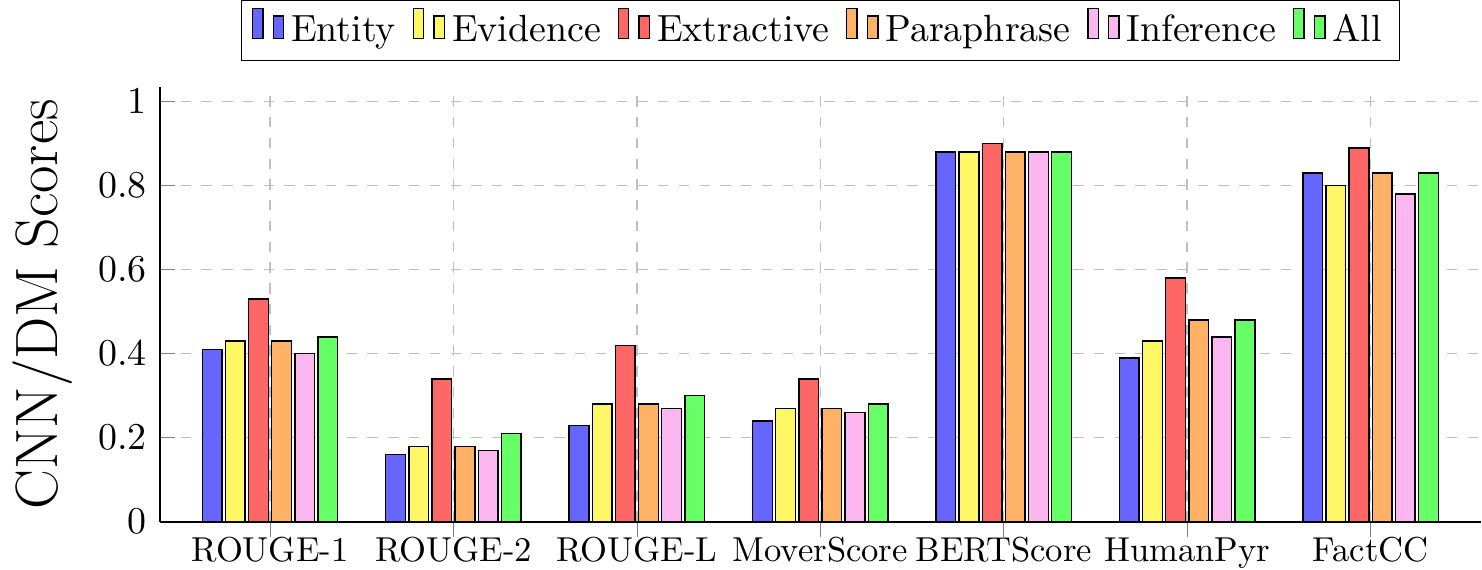}
    \caption{CNN/DM class-level performance, averaged across all models.}
    \label{fig:classcnndm}
\end{figure}

\begin{figure}[t]
    \centering
    \includegraphics[width=1\columnwidth]{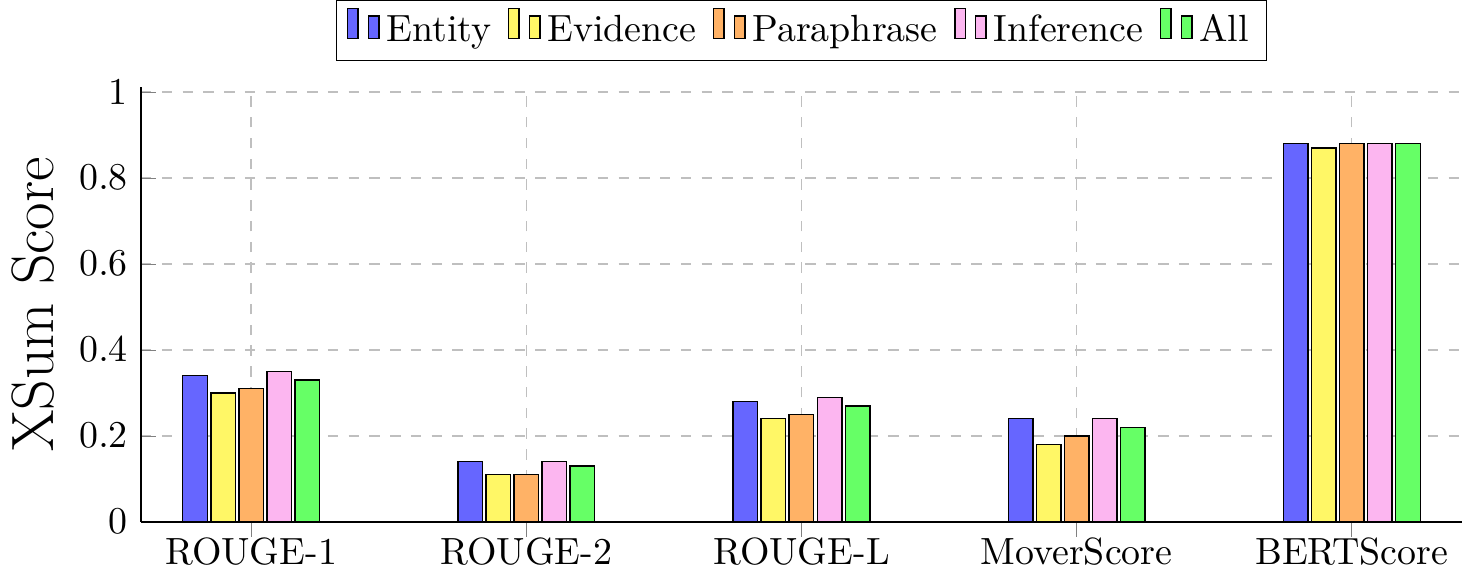}
    \caption{XSum class-level performance, averaged across all models.}
    \label{fig:classxsum}
\end{figure}

For each dataset, we group the samples by their labels. For all samples in a subset, the model response is scored using a metric. The mean of these sample scores returns a single subset-model-metric score, which is then averaged across all models in the subset, leaving us with a single subset-metric score. This is repeated for all (subset $\times$ metric) pairs. The results are captured in Figures \ref{fig:classgiga}, \ref{fig:classcnndm} and \ref{fig:classxsum} for Gigaword, CNN/DM and XSum respectively. The last column in each group is the average score across all samples.

\subsubsection{Impact of Data Quality and Noise }

\paragraph{Incomplete and Irrelevant} Of the three datasets, only Gigaword contains Incomplete (or Irrelevant) samples. Across all metrics, the performance on this label is lowest, which is to be expected -- high overlap will be rare if the source and target are unrelated or incomplete (like Row 1,  Table \ref{tab:datasamples}). What's alarming is the volume of such samples in Gigaword -- if the distribution is the same for the training set, then the model is being trained on extremely noisy data (almost 14\%). In addition, such samples needlessly penalise the model performance during evaluation.

\paragraph{Entity scores more than Evidence in Gigaword!} The results for these subsets are a bit surprising. In Gigaword, the Entity Missing subset receives relatively higher scores than the Evidence Missing category. We attribute this to a combination of factors. Consider Row 2 in Table \ref{tab:datasamples}. Entities are missing, but token overlap is high (more than 50\%), which explains the high R1 scores, but low R2 scores. In our observations, the impact of missing facts and entities varies by the length of the target, as well as the number of entities.

\paragraph{Are Evidence Missing and Paraphrase are all the same for CNN/DM and XSum?} When compared with Gigaword, samples with data quality issues (i.e. Incomplete/Irrelevant, Entity Missing and Evidence Missing samples) in CNN/DM and XSum get relatively higher scores. The reasons are similar to the Gigaword phenomenon discussed before. The average summary length of CNN/DM (54 tokens) is about 7 times that of Gigaword (8 tokens). As a result, with respect to the complete reference, one or two missing facts amounts to a much smaller fraction of the reference in CNN/DM. The high overlap with the remainder leads to higher scores.

\paragraph{Factual Correctness in CNN/DM} Automatic metrics only consider the token overlap (or ``semantic distance") between the target and the model output. While such metrics exhibit high correlation with human-judgement, a low score does not necessarily imply an incorrect generation, as demonstrated by \citet{freitag2020bleu} for machine translation. Hence we check for factual correctness of model outputs using FactCC. The competitive scores on the first three categories for FactCC in Fig~.\ref{fig:classcnndm} suggests the outputs generated by the model are factually faithful, which points to issues with the metric reliability. We discuss this in Section \ref{section:q2b}.

\subsubsection{Impact of Summarization Complexity}

For the last three categories (Extractive, Paraphrase and Inference) Gigaword and CNN/DM exhibit a common trend: the highest performance, across all metrics is on the Extractive subset, followed by Paraphrase samples which are more difficult to reproduce. The lowest performance is on the Inference samples. However, concluding models perform poorly would be incorrect. The last three samples in Table \ref{tab:datasamples} suggest that model outputs are coherent, logical and factually faithful. FactCC scores in Figure \ref{fig:classcnndm} also suggest the outputs are factually consistent.

\paragraph{Some metrics are biased towards simpler samples?}
For the Extractive, Paraphrase and Inference samples, the samples we manually observed (some of which are captured in Table \ref{tab:datasamples}) and the FactCC scores indicates a gap in the token-based metrics. However, we cannot fault the metrics entirely. If we had diverse target references for the same sources, some outputs would have found better matches, and thus, higher scores! In fact, we see that BERTScore (a more ``semantically" oriented metric) is extremely competitive across all categories in all three datasets (Figures \ref{fig:classgiga}, \ref{fig:classcnndm}, \ref{fig:classxsum}), suggesting the generations are similar to the references. These results lead us to believe that token-based summarization metrics might also suffer from a ``summarization-ese" effect: \textbf{the metrics could be biased towards simpler, more ``extractive" references}. Recently, \citet{freitag2020bleu} also arrived at the same conclusion for machine translation and BLEU \cite{papineni-etal-2002-bleu}.

In the next section, we continue to explore the reliability of these metrics.

\section{Does the reliability of metrics change with data properties? (Q2 b)} 
\label{section:q2b}
For each document $d_i, i \in \{1 \dots n\}$ in a dataset $\mathcal{D}$, we have $J$ system outputs, where the outputs can come from different systems. Let $s_{ij}, j \in \{1 \dots J\}$ be the $j^{th}$ summary of the $i^{th}$ document, $m_i$ be a specific metric (including human judgment).

\begin{equation}
    \begin{split}
        K_{m_1 m_2}^{sum} &= \frac{1}{n} \sum_{i = 1}^{n} \bigg( K\big([m_1(s_{i1}) \dots m_1(s_{iJ})], \\
        & ~~~~~~~~~~~~~~~~~~~~~~~~[m_2(s_{i1}) \dots m_2(s_{iJ}) ]\big) \bigg).
    \end{split}
    \label{eqn:summ_level_corr}
\end{equation}

Correlation is calculated for each document, among the different system outputs of that document, and the mean value is reported. Like other meta-evaluation studies, we consider the Pearson correlation and Spearman correlation as measures for $K$. Due to space constraints we only show the Pearson plots for some critical results. More plots are available in Appendix \ref{section:appx-figs}.




\begin{figure}[ht]
  \centering
  \includegraphics[width=1\columnwidth]{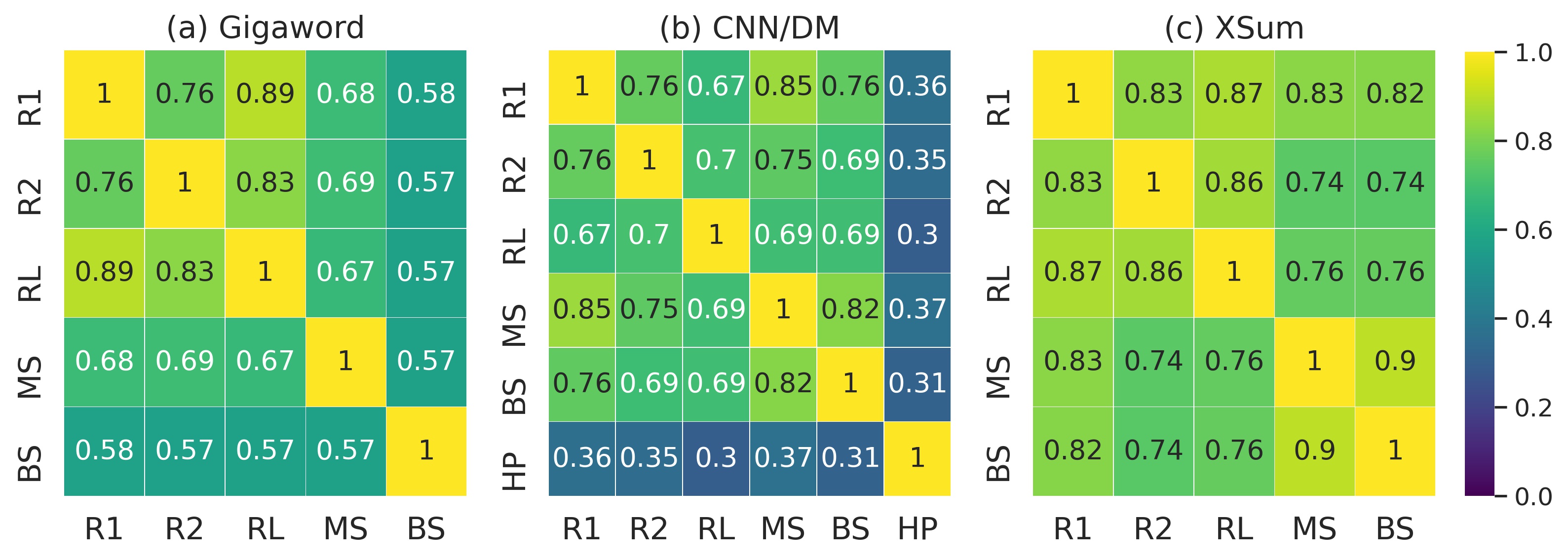}
     \caption{Pearson correlation between different metrics for all three datasets.}
     \label{fig:unified_pearson}
\end{figure}

\paragraph{Inter-metric Correlation}
We present a pairwise correlation analysis of the automatic metrics to understand metric agreement in Figure \ref{fig:unified_pearson}. We conjecture that a strong correlation between two vastly different metrics (say ROUGE and MoverScore) might show that the metric is more reliable. Overall, we can see in Figure \ref{fig:unified_pearson} that correlations between token-based metrics (ROUGE) and embedding-distance metrics (BERTScore, MoverScore) is lower in Gigaword, compared to CNN/DM and XSum. It is possible that the short length summaries of Gigaword is leading to this; perhaps there isn't enough context for BERTScore. Although, we could not find any results in the original papers to support this claim.



\begin{figure}[t]
    \centering
    
    \subfloat[Gigaword]{
    \includegraphics[width=1\columnwidth]{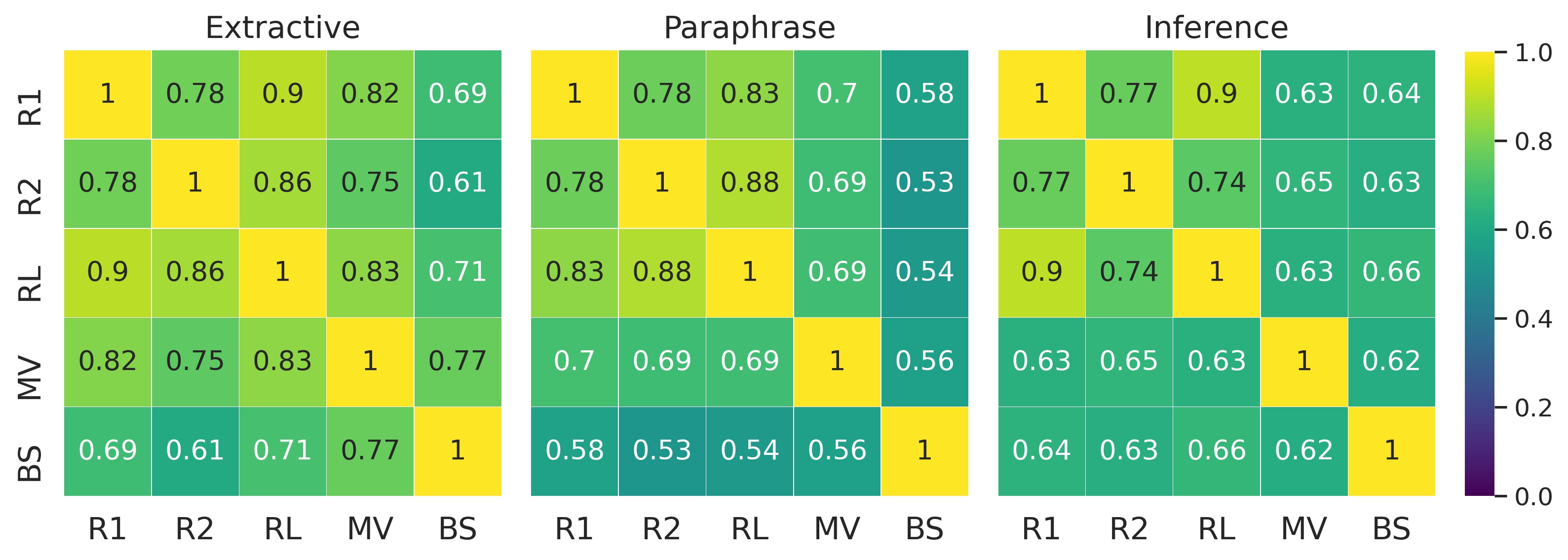}
    }  \\
    \subfloat[CNN/DM]{
  \includegraphics[width=1\columnwidth]{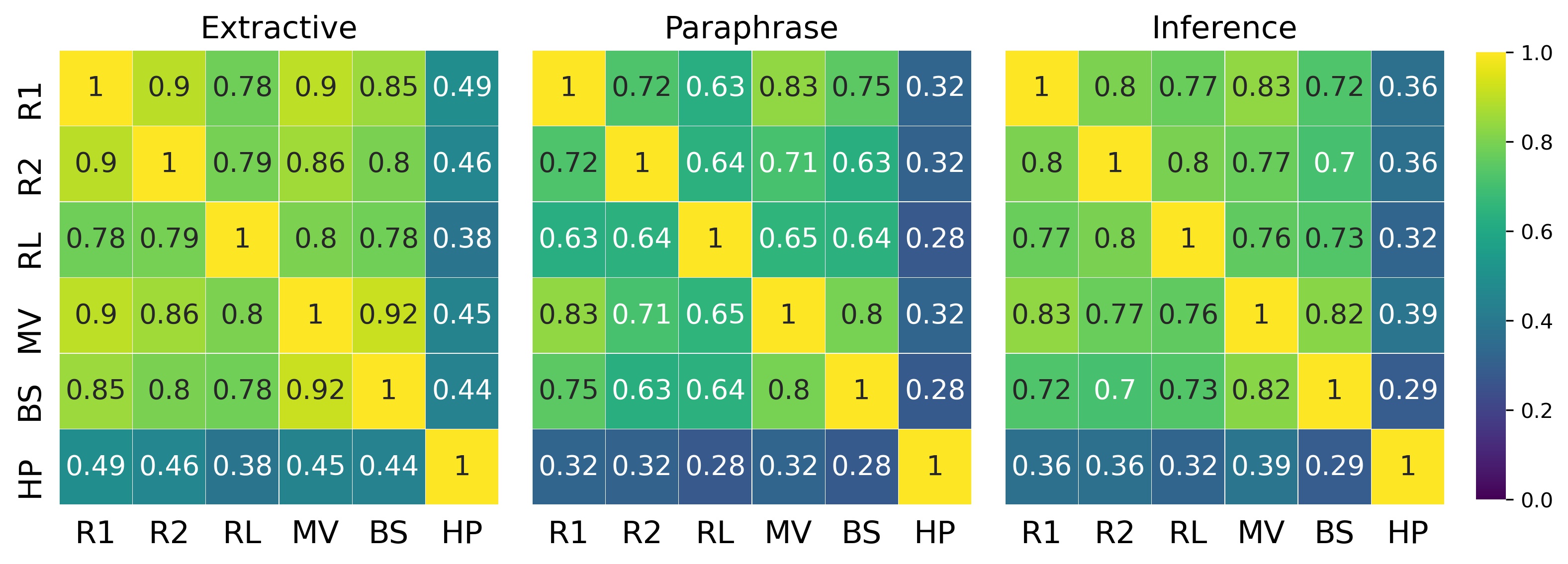} 
    }
    \caption{Pearson correlations for Extractive, Paraphrase and Evidence samples in Gigaword and CNN/DM.}
    \label{fig:pcorrelation}
\end{figure}

\paragraph{Correlation variation with complexity} We observe that the correlation is heavily sample dependent. In Figure \ref{fig:unified_pearson}, averaged across all samples, R1 and MoverScore have a Pearson correlation of about 0.68 in Gigaword. This increases to 0.82 for the Extractive samples in Figure \ref{fig:pcorrelation}-(a), which are the simplest to reproduce. As the complexity increases, the correlation scores decrease (in Paraphrase, and then in Inference). The trends for R2 and MoverScore are similar. This is also observed for CNN/DM: in Figure \ref{fig:pcorrelation}-(b), correlations for R1-MoverScore and R1-BERTScore drop from 0.9, 0.85 for Extractive samples to about 0.83, 0.72 for Paraphrase and Inference samples. \textbf{This suggests that the inter-metric correlation is heavily sample dependent}. We cannot comment on XSum, because we did not encounter any Extractive samples in that dataset.

\paragraph{Correlation with Human Judgement}
For CNN/DM, we also compute the metric correlations with the human pyramid score (HP) in Figure \ref{fig:unified_pearson} and Figure \ref{fig:pcorrelation}-(b). We observe the highest agreement with the human-judgement for the Extractive subset, and it is significantly lower in Paraphrase and Inference.  \textbf{This suggests that automatic metrics are more reliable when evaluating simpler examples, than complex ones}.


\section{Discussion}
\label{section:discussion}

\noindent\textbf{Limitations of the typology.} Forcing samples to have a single label did limit our analysis. In retrospect, the typology could have allowed for two labels: one for quality, one for complexity. In XSum for instance most samples which were labelled Entity Missing could also be labelled Paraphrase and Inference. We also realise that the impact of positional-bias could be important. This has been explored by \citet{zhong-etal-2019-searching, zhong-etal-2019-closer}, and we plan to include similar metrics in our future work.

\noindent\textbf{Collecting better datasets.} Our results suggest that current metrics are not equally reliable across all categories of samples. If the quality of the references cannot be controlled, then having a diverse set of references for the source is also advised. This will allow for multi-reference evaluation and could offset the ``summarization-ese" issues.

\noindent\textbf{Limits of the Pyramid Scores.} At the moment, the Pyramid Scores (and judgement criteria in general) only compare the output to the gold-reference, assuming the latter is \textit{true}. As we see from our analysis, ignoring the source is not the right approach, for references from the web could have quality issues. A modified judgement procedure, that also accounts for the faithfulness of the gold-reference (perhaps by using automatic factuality metrics FactCC) might be better.

\noindent\textbf{Architecture specific performance.} In this study, we were interested in measuring the broader, \textit{averaged} trends that summarization models exhibit. However, it would be interesting to see how specific architectural decisions impact individual model performance across different classes. We plan to explore this in the future.

\noindent\textbf{``But what's the best metric for my data?"} Specifically for metrics, our objective was to empirically demonstrate that (a) datasets have different modalities, and (b) metrics are not equally reliable across these modalities. In this process, we also observed some results suggesting possible biases in certain token-based metrics, and a need for diverse reference sets. We'll continue to explore this question.

\section{Related Work}

For the task of text-summarization, the data analysis heuristics presented in \citet{zhong-etal-2019-searching, zhong-etal-2019-closer, bommasani-cardie-2020-intrinsic, grusky-etal-2018-newsroom} are most relevant to our work. Their analysis is focused on surface level heuristics which ignores all noise present in the data. This has been discussed in Sections \ref{section:motivate}, \ref{section:discussion}. Researchers have also explored other dataset biases~\cite{jung-etal-2019-earlier, zhong-etal-2019-closer,chen-etal-2020-cdevalsumm}. As discussed in Section \ref{section:discussion}, we plan to include this in our future work. 

For metric reliability and meta-analysis, we build on correlation analysis presented in earlier works \cite{peyrard-2019-studying, bhandari-etal-2020-evaluating, fabbri2020summeval}. The key difference and novelty is the introduction of our typology and measuring the impact of sample complexity on model performance and metric reliability. To the best of our knowledge, metrics and models have not been evaluated on such a typology. As results in Section \ref{section:q2a} and \ref{section:q2b} show, sample complexity is indeed very critical for metric reliability.




\section{Conclusion}
\label{section:conclusion}
In this study, we manually analysed 600 samples from three popular datasets, using a typology that captures data quality issues and varying degrees of sample-complexity. Our analysis of 27 summarization models reveals that the metric performance is heavily dependent on samples. On closer inspection, we found that the agreement of popular metrics also changes with the complexity, thus the scores might not reflect true model performance. This analysis also led to some suggestions for creating better summarization datasets and highlights some limitations of the current human-judgement procedures.

\section*{Acknowledgements}

We thank Professor Graham Neubig, Yiran Chen and anonymous reviewers for valuable feedback and helpful suggestions.
Thanks Kaiqiang Song for providing system outputs.
This work was supported in part by a grant under the Northrop Grumman SOTERIA project and the Air Force Research Laboratory under agreement number FA8750-19-2-0200. 
\bibliography{acl2020}
\bibliographystyle{acl_natbib}

\clearpage

\FloatBarrier

\begin{appendices}

\section{Figures and Annotation Details}
\subsection{Correlation plots}
\label{section:appx-figs}



\begin{figure}[!htb]
    \centering
    \subfloat[Pearson]{
    \includegraphics[width=1\columnwidth]{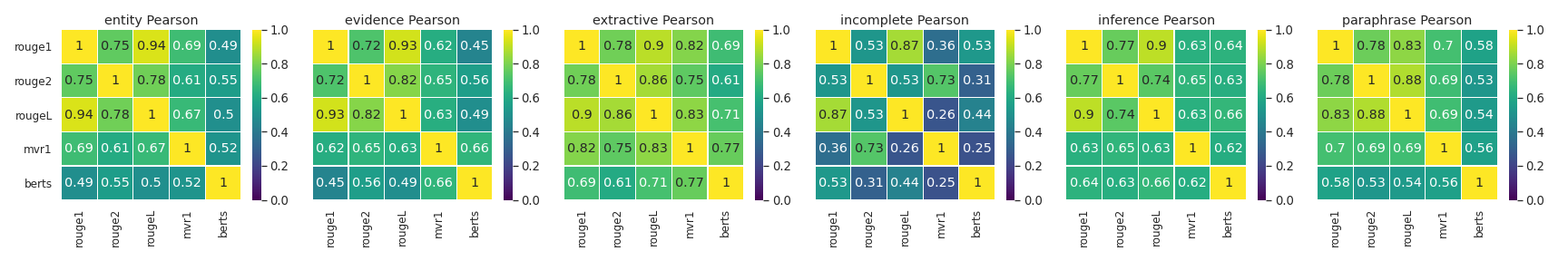}
    } \\
    \subfloat[Spearman]{
  \includegraphics[width=1\columnwidth]{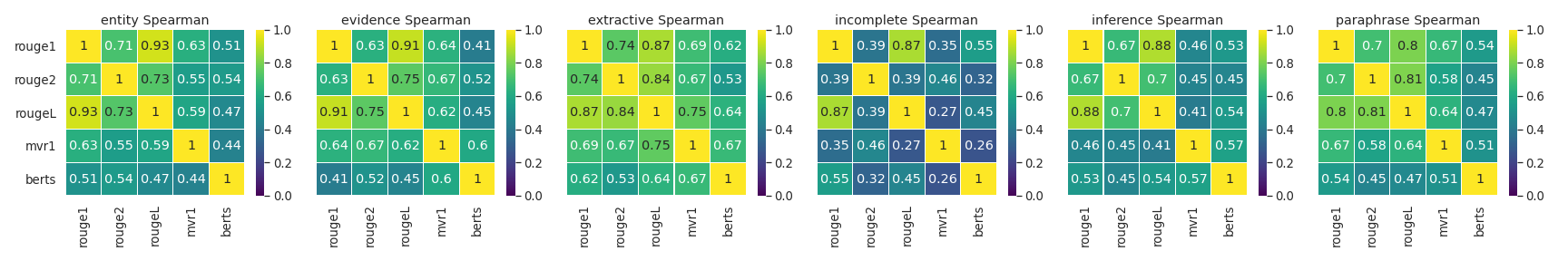} 
    }
    \caption{Gigaword correlations.}
    \label{fig:scorrelation-giga-appx}
\end{figure}



\begin{figure}[!htb]
    \centering
    \subfloat[Pearson]{
    \includegraphics[width=1\columnwidth]{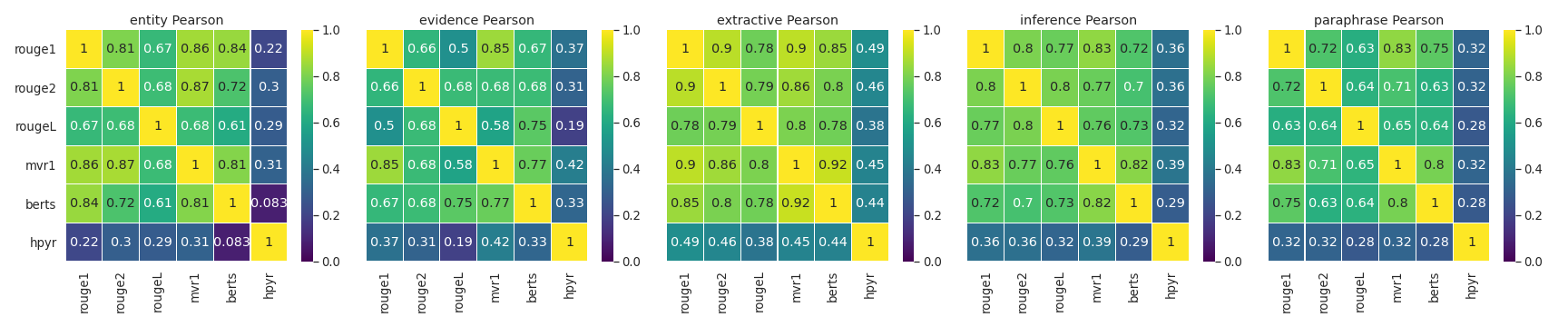}
    } \\
    \subfloat[Spearman]{
  \includegraphics[width=1\columnwidth]{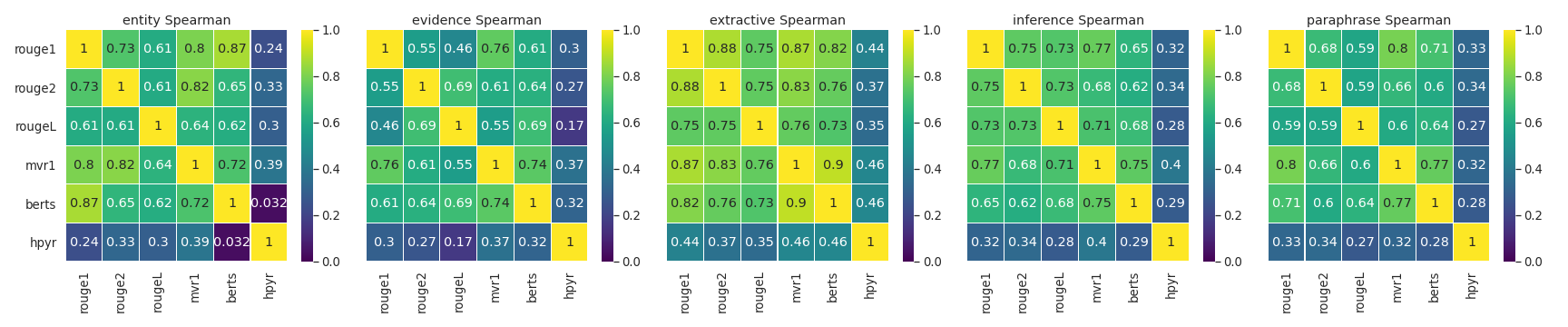} 
    }
    \caption{CNN/DM correlations.}
    \label{fig:scorrelation-cnn-dm-appx}
\end{figure}



\begin{figure}[!htb]
    \centering
    \subfloat[Pearson]{
    \includegraphics[width=1\columnwidth]{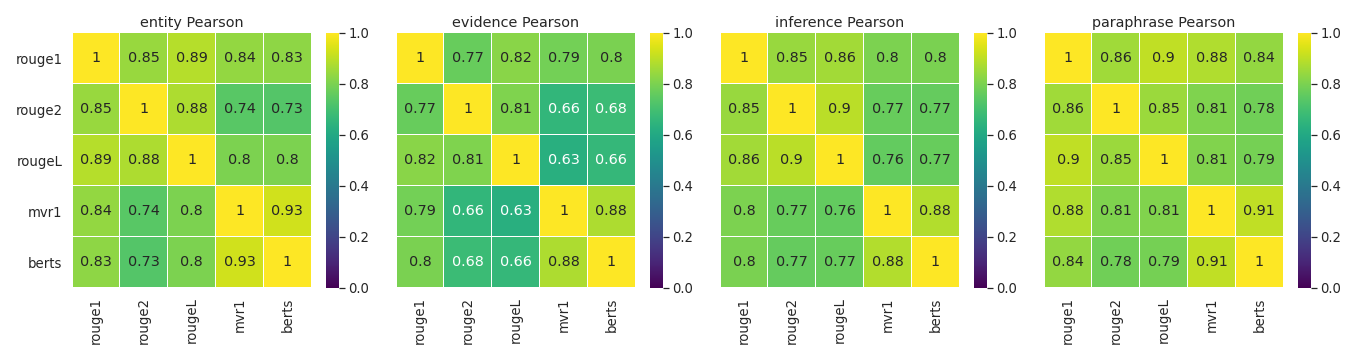}
    } \\
    \subfloat[Spearman]{
  \includegraphics[width=1\columnwidth]{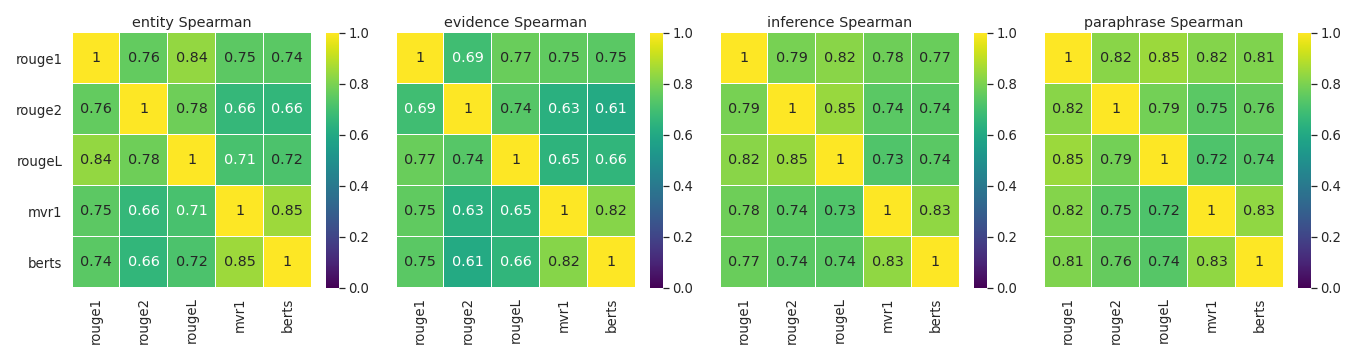} 
    }
    \caption{XSum correlations.}
    \label{fig:scorrelation-xsum-appx}
\end{figure}

\subsection{Annotation Details}
\label{section:appx-anno}
Each sample is annotated by 2-3 annotators independently. Given the limited number of samples, and the laborious nature of the exercise, we chose not to select final labels based on majority vote. For all disagreements, annotators discussed their reasoning and came to an consensus for final label. For 70\% of Gigaword samples, 68\% of CNN-DM samples, and 73\% of XSum samples, the initial annotations were in agreement.

\end{appendices}
\FloatBarrier

\begin{appendices}
\section{Gigaword}
\label{section:appx-giga}
\subsection{Gigaword: Paraphrase and Inference samples}




\begin{table}[htbp]
  \centering
  \tiny
    \begin{tabular}{ p{1cm}  p{6cm} }
    \toprule
    \textbf{Label} & \textbf{Source} \\ \cmidrule{2-2} 
        & \textbf{SoTA Output} \\ \cmidrule{2-2} 
        & \textbf{Gold Reference} \\
        \midrule
        
    Paraphrase & A woman street cleaner and her three young daughters were killed Saturday when a bomb in a metal container exploded in Bangladesh , police said . \\ \cmidrule{2-2}
        & Mother , three daughters die in in Bangladesh blast . \\ \cmidrule{2-2}
        & Mother , three daughters killed in Bangladesh blast .
        \\
        \midrule
        
    Paraphrase &
        The UN chief of Eastern Slavonia , the last Serb-held part of Croatia , confirmed Tuesday that key elections would be held here on April 13 as part of local ballots throughout Croatia . \\ \cmidrule{2-2}
        & UN chief confirms key elections in Eastern Slavonia . \\ \cmidrule{2-2}
        & UN confirms elections to be on April 13 in Eastern Slavonia .
        \\
        \midrule
        
    Paraphrase &
        Business at Taiwan 's theme parks and resorts grew significantly in the first quarter of this year compared to Q1 last year , the Tourism Bureau said Thursday , attributing the growth to the government 's shopping voucher program and other promotion efforts . \\ \cmidrule{2-2}
        & Business at Taiwan 's theme parks and resorts grows . \\ \cmidrule{2-2}
        & Shopping vouchers help boost theme parks business : tourism bureau .
        \\
        \midrule
    
    Inference &
        Col. Robert E. Lee skirted the unleaded gasoline pit , negotiated a thicket of telephone cords stretched as tight as trip wires and took the center of the New York Mercantile Exchange 's main trading floor just before 3 p.m. last Monday . \\ \cmidrule{2-2}
        & New York Mercantile Exchange 's trading floor . \\ \cmidrule{2-2}
        & MILITARY STRATEGISTS PRACTICE IN REAL BATTLE ON WALL STREET .
        \\
        \midrule
        
    Inference &
        Finland scored three goals in a 40-second span of the first period Tuesday night for a 7-3 victory over the Czech Republic in their World Cup of Hockey opener . \\ \cmidrule{2-2}
         & Finland 7 , Czech Republic 3 . \\ \cmidrule{2-2}
         & Finland Routs Czech Republic at World Cup .
        \\
        \midrule
        
    Inference &
        Q. I 've heard that cow manure can be used for energy production , but not human waste . \\ \cmidrule{2-2}
         & Cow manure can be used for energy production . \\ \cmidrule{2-2}
         & ON NOT WASTING WASTE .
        \\
        \midrule
        
    \bottomrule
    \end{tabular}%
    
    \caption{Source, outputs and targets, from Gigaword.}
    \label{tab:appendixgiga}%
\end{table}%

\FloatBarrier

\end{appendices}
\FloatBarrier

\begin{appendices}

\section{CNN/DM}
\label{section:appx-cnndm}

\subsection{CNN/DM: Paraphrase and Inference samples}

\begin{table*}[!htb]
  \centering
  \tiny
    \begin{tabular}{ p{1cm}  p{14cm} }
    \toprule
    \textbf{Label} & \textbf{Source} \\ \cmidrule{2-2} 
        & \textbf{SoTA Output} \\ \cmidrule{2-2} 
        & \textbf{Gold Reference} \\ \midrule

    Paraphrase &
    
        \textcolor{blue}{Her neighbour’s leylandii hedge stands 40ft tall and, says Audrey Alexander, has left parts of her garden in deep shade.} What’s more, it now seems likely to remain that way. (truncated). A row between neighbours over a 40ft high leylandii hedge (pictured) has finally come to and end after 35 years . \textcolor{blue}{The battle between the neighbouring properties started in 1980 when the owner planted a vegetable patch which withered and died in the shade of her neighbour's massive hedge . Then, 23 years ago, single mother Mrs Alexander bought the house and asked her neighbour Jeanette Robinson to trim the hedge.} She claims Mrs Robinson refused and declared: ‘I would rather move than touch these trees.’ (truncated) . Audrey Alexander (pictured) also claims other neighbours have had to move their children from their bedrooms at night for fear of the falling branches . \textcolor{blue}{But her council has ruled that Mrs Robinson can keep the hedge, although it has to be cut to 20ft.} Mrs Alexander said the ruling made ‘no difference’. (truncated) 
        \\ \cmidrule{2-2}
        & Audrey Alexander 's vegetable patch withered and died in the shade of hedge .  She asked neighbour Jeanette Robinson to trim it but she refused .  Mrs Alexander claims hedge knocked £ 20,000 off the value of her house .  Stirling Council has ruled that Mrs Robinson can keep the hedge .  But it has to be cut to 20 ft , a height which she claims will still block most of her sunlight .
        \\ \cmidrule{2-2}
        & Audrey Alexander wanted her neighbours to chop down their huge hedge .  She claims the 40 ft leylandii was blocking sunlight from reaching her home .  Feud started in 1980 when it blocked light from reaching a vegetable patch .  Council finally rules that the hedge can stay - but must be cut back to 20 ft   .

        \\
        \midrule

    Paraphrase & 
        \textcolor{blue}{The number of women in Britain becoming nuns is at a 25-year high. Figures from the Catholic Church show the number of women taking Holy Vows has trebled from 15 in 2009 to 45 last year. From a low of only seven in 2004, the figure has been rising for the past decade.} Theodora Hawksley, 29, was until recently a post-doctoral researcher in theology at the University of Edinburgh. (truncate). Far from being trapped in traditional habits, Miss Hawksley said her order tends to dress down in T-shirts and jeans. \textcolor{blue}{Father Christopher Jamison, director of the National Office for Vocation of England and Wales, said: ‘There is a gap in the market for meaning in our culture. One of the ways women may find that meaning is through religious life.’} Sister Cathy Jones, religious life vocations promoter at the office, said: (truncated) .
        \\
         \cmidrule{2-2}
         
        & Figures from Catholic Church show number of women taking Holy Vows has trebled from 15 in 2009 to 45 last year .  From a low of seven in 2004 , the figure has been rising for the past decade .  Theodora Hawksley , 29 , was until recently a post - doctoral researcher in theology at the University of Edinburgh .  But at the beginning of the year she decided to become a nun .
        \\ \cmidrule{2-2}
        & Figures from the Catholic Church show more and more becoming nuns .  The number of women taking Holy Vows stood at just seven back in 2004   .  But that figure had risen to 15 in 2009 and increased further to 45 last year .  One father said a ' gap in the market for meaning ' led people toward religion .
     
        \\
        \midrule
        
    Inference &
        \textcolor{blue}{Following all his inspired charity work, Didier Drogba has been awarded with a Barclays Spirit of the Game trophy. The Chelsea forward set up the 'Didier Drogba Foundation in Africa,' as he hopes to inspire the next generation of footballers in Africa to fall in love with the game.} (truncated) He said 'I come from a poor family where I played football in the streets with my friends with no shoes, there was no grass but we still enjoyed it. \textcolor{blue}{The 'Didier Drogba Foundation,' contribute financial and material support in education and health including school bags for the school children, as well as a medical clinic in his hometown of Abidjan, Ivory Coast, which will be opening its doors later this year. Chelsea's stars such as Eden Hazard, Petr Cech and Branislav Ivanovic were out in force earlier this month as they raises £400,000 for the foundation at a charity ball.} The money raised will be used to complete the medical clinic in Abidjan and help finance mobile clinics that will travel outside of the capital to those who are either to sick or poor to make the journey to the medical centre.
        \\ \cmidrule{2-2}
        
        & Didier Drogba has been awarded with a Barclays Spirit of the Game trophy .  The Chelsea forward set up the ' DidierDrogba Foundation in Africa ' He hopes to inspire the next generation of footballers in Africa to fall in love with the game .  The 37-year - old scored the equaliser against Leicester on Wednesday .
        \\ \cmidrule{2-2}
     
        & Didier Drogba given the Barclays Spirit of the Game award .  The 37-year - old 's foundation has done impressive work in Africa .  Some of Chelsea 's stars attended a charity ball which raised £ 400,000 .  CLICK HERE for all the latest Chelsea news   .
     
        \\
        \midrule

    Inference &
        (truncated) Resorts on its Black Sea coast offer the best value in terms of a meal out, buying a cup of coffee and essentials such as sun cream and a cold drink, according to a study. Scroll down for video . \textcolor{blue}{Affordable: Bulgaria has been named Europe's cheapest destination, with Black Sea resorts like Sunny Beach (pictured) offering the best value in terms of a meal out and other holiday activities .} Hotspot: \textcolor{blue}{Bulgaria's most popular resort of Sunny Beach is a carbon copy of those of Spain and Greece} . \textcolor{blue}{It is one of 13 European hotspots out of 14 where your cash will go far further this summer, largely thanks to rock-bottom exchange rates and higher inflation in some countries. Research into an imaginary shopping basket of ten typical holiday purchases showed a total price of £37.39 for Bulgaria, which is down by 13.6 per cent from last summer. There was a bigger fall of 22 per cent for the Algarve in Portugal, taking the total cost to £44.02, helping it beat Spain's Costa del Sol to become the second cheapest destination. Only in Turkey, where inflation is 7.6 per cent – compared to virtually zero in Britain and the eurozone – will Britons find the cost of a day out much more expensive. The figures, compiled for the annual Post Office Holiday Costs Barometer, show the spending basket in Turkey is up by 21.4 per cent on last year, at £65.70.} Bulgaria's most popular resort of (truncated) .
        \\
        \cmidrule{2-2}
     
        & Former Soviet state has gained the most from the strong pound .  Resorts on its Black Sea coast offer the best value in terms of a meal out , buying a cup of coffee and essentials such as sun cream and a cold drink .  It is one of 13 European hotspots out of 14 where your cash will go far further this summer .
        \\
        \cmidrule{2-2}
        
        & Bulgaria's Black Sea resorts cheaper than hotspots in Italy, Spain and Turkey . Researchers found cheapest destination using 'imaginary shopping basket' Cheap prices are driven by low exchange rates and country's high inflation . Its most popular resort of Sunny Beach copies those of Spain and Greece .
        \\
        \cmidrule{2-2}

    \bottomrule
    \end{tabular}%
    
    \caption{Source, outputs and targets, from CNN/DM.}
    \label{tab:appendixcnndm_paraphrase}%
\end{table*}%

\end{appendices}
\FloatBarrier

\begin{appendices}

\section{XSum}
\label{section:appx-xsum}

\subsection{XSum: Paraphrase and Inference samples}

\begin{table*}[htbp]
  \centering
  \tiny
    \begin{tabular}{ p{1cm}  p{14cm} }
    \toprule
    \textbf{Label} & \textbf{Source} \\ \cmidrule{2-2} 
        & \textbf{SoTA Output} \\ \cmidrule{2-2} 
        & \textbf{Gold Reference} \\
        \midrule

    Paraphrase &
        \textcolor{blue}{More than 700,000 employees face unpaid leave due to the shutdown which was triggered after the two houses of Congress did not agree on a new budget. Hyundai said affected employees who currently own its vehicles will be given a payment relief "for as long as they are out of work".} Employees looking to buy a new car will be given a 90-day payment deferral. "We recognize the impact on family budgets that the furlough will drive," John Krafcik, chief executive of Hyundai Motor America, said in a statement. Hyundai had offered a similar scheme, the Hyundai Assurance programme, during the peak of the global financial crisis four years ago to help consumers who had lost their jobs. Many analysts have said that the move had helped the South Korean firm win customer loyalty and boosted its sales in recent years. The company said that its latest offer to help the federal employees was an addition to that programme and aimed at "helping workers at a time when they most need it". "Like we did almost four years ago when we launched Hyundai Assurance, this is our way of saying 'We've got your back' during this uncertain time," Mr Krafcik said. Under the latest offer, Hyundai will extend all auto loan and lease payments during the shutdown for current Hyundai owners who are put on unpaid leave. The programme is available to all customers who have financed their purchase or lease through Hyundai Finance America.
        \\ \cmidrule{2-2}
        & US carmaker Hyundai Motor has offered financial help to federal employees who have been affected by the government shutdown . \\ \cmidrule{2-2}
        & Hyundai Motor will defer payments due from US federal employees affected by the partial government shutdown .
        \\
        \midrule

    Paraphrase &
        \textcolor{blue}{Gary Price was suspended from all council duties for five months in November after Powys council's Standards Committee ruled he had breached the code of conduct. His appeal has been dismissed by the Adjudication Panel for Wales following a two-day hearing in Llandrindod Wells.} Mr Price has been contacted for comment. He was found to have sent information which the council said "incorrectly and unfairly" portrayed what happened at a grievance appeal hearing, in which he was a panel member. The Adjudication Panel for Wales unanimously agreed to refer the matter back to the Standards Committee with a recommendation that Mr Price be suspended for three months. Council leader Barry Thomas said the decision "sends out a clear message that those who enter public office have to operate within the members' code of conduct and maintain the highest possible standards".
        \\ \cmidrule{2-2}
        & A Powys council chief executive has lost his appeal against a decision to suspend him .
        \\ \cmidrule{2-2}
        & A decision to suspend a Powys county councillor has been upheld .
        \\
        \midrule

    Inference &
        \textcolor{blue}{Derby City Council wanted to shut Moorways Pool from April in a bid to save about Â£350,000 a year. The Labour-led authority, which needs to save Â£79m over the next three years, said it had found the savings by making cuts in other areas}. Campaigners who gathered more than 4,000 signatures on a petition said they were delighted at the news. \textcolor{blue}{Ranjit Banwait, leader of the authority, said the council had committed to keep it open for a year}. He said the council had identified savings "in back-office areas" and a restructuring of management jobs, which had been "untouched" since 2010. However, he stressed if the authority failed to get a "fair deal" from central government in the future, the pool would still have to close. Campaigners had accepted the pool, which is 33m in length, was in need of repair. There are plans for a new 50m pool to be built by 2018 to replace it. However, closing it would have left only one other public pool in the city - the Queen's Leisure Centre, they said. Doug Whitlam, of the Derbyshire Amateur Swimming Association, said: "One of the main things for me would have been the loss of teaching. "Twelve hundred young people use this facility every week and that would be lost forever."
        \\ \cmidrule{2-2}
        
        & A council has backed down over plans to close a public swimming pool in a bid to save money .
        \\ \cmidrule{2-2}
        & A Derby swimming pool threatened with closure is to remain open for another year , council bosses have confirmed .
        \\ \midrule

    Inference &
        It is likely to include a scrappage scheme for older diesel cars in areas with high levels of dirty air. Speed bumps could be removed in some cities to cut pollution from cars slowing down and speeding up. \textcolor{blue}{Environmental lawyers ClientEarth said they would "thoroughly analyse" the proposals.} According to the Royal College of Physicians, air pollution across the UK is linked to around 40,000 premature deaths every year. The UK has struggled to keep within EU limits on some pollutants, particularly nitrogen dioxide (NO2), which is produced by diesel engines and is linked to a range of respiratory diseases including asthma. Some 37 of the 43 regions of the UK are in breach of NO2 limits. Under earlier government plans, some parts of the UK would not have met EU NO2 standards until 2030. \textcolor{blue}{The original deadline to achieve these limits was 2010. Exasperated by what they believed was government foot-dragging on the question of cleaner air, ClientEarth mounted a legal challenge to force faster action. In April 2015, the UK Supreme Court ruled the government had to take immediate steps on the issue. Unhappy with the timescales in the plan that was then produced, ClientEarth went to the High Court last November for a judicial review. Once again the court supported the lawyers, telling the government that its scheme was "woefully inadequate" and giving ministers until 24 April this year to produce a new draft.} With a general election in the offing, the government last week asked the judge for permission to delay the draft plan. \textcolor{blue}{But Mr Justice Garnham disagreed and ordered publication by 9 May. "These steps are necessary in order to safeguard public health," he said. Earlier this week, the government said it would not appeal against the ruling and would publish}. In their previous plans, ministers wanted to create "clean air zones" in five cities outside London with high levels of NO2. Only the most polluting vehicles would have to pay a charge to enter the zone under that scheme. The new draft plan is expected to create many more such zones. Councils will be given the power to impose fines or restrictions on all polluting vehicles in these areas. In the worst cities, so called "toxin taxes" could range up to Â£20 a day but the government is said to be keen not to punish drivers who bought diesels as a result of incentives brought in by a previous Labour administration. This is something that the lawyers at ClientEarth support. "Successive governments have encouraged people to buy diesel. We don't want to see diesel drivers vilified, and we think the plans should also include properly funded incentives to help people move to cleaner forms of transport," said ClientEarth CEO James Thornton. "We will thoroughly analyse the government's draft plans when they are produced. If we do not think they are in line with the court order, to deal with illegal levels of pollution as soon as possible, then we will consider our next steps." According to newspaper reports, the government has agreed to back a "targeted" scrappage scheme for older diesel cars, but limited to vehicles in areas of high pollution. There may also be funding for a retrofitting scheme to help existing diesel car and van owners cut their emissions of NO2. The government is also said to be pushing for councils to use alternatives to charging, including the removal of speed bumps in some places and the better sequencing of traffic lights in others. Both of these measures could limit cars having to slow down and speed up repeatedly, actions that can almost double the amount of NO2 produced. However, the idea that speed bumps which slow down traffic would be sacrificed to help clean up the air we breathe is not a welcome concept according to road safety charity Brake. "We ought not to be made to choose between having cleaner air and safer roads," a spokesman said. "The evidence shows that air pollution is contributing to the early deaths of thousands of people. It's now clear that there's more than one way a car can kill you." The new proposals will be out for consultation for six weeks before the government produces a final plan at the end of July. Follow Matt on Twitter and on Facebook.
        \\ \cmidrule{2-2}
        
        & The government is expected to publish a new draft plan to tackle air pollution in the UK later this week .
        \\ \cmidrule{2-2}
        
        & The UK government is set to publish a draft air pollution plan after a protracted legal battle with environmental campaigners .
        \\
        \midrule

    \bottomrule
    \end{tabular}%
    
    \caption{Source, outputs and targets, from XSum.}
    \label{tab:appendixgiga}%
\end{table*}%

\end{appendices}

\end{document}